\documentclass[a4paper,10pt]{article}
\usepackage{geometry}
\usepackage[utf8]{inputenc} 
\usepackage[T1]{fontenc}    
\usepackage{hyperref}       
\usepackage{url}            
\usepackage{booktabs}       
\usepackage{amsfonts}       
\usepackage{nicefrac}       
\usepackage{microtype}      
\usepackage{xcolor}         

\usepackage{lipsum}
\usepackage{graphicx}
\usepackage{amsmath,amsfonts,bm}
\newcommand{\vx}{\bm{x}}

\title{Kolmogorov-Arnold Networks are \\ Radial Basis Function Networks}

\author{
    Ziyao Li \\
    Center for Data Science, Peking University \\
    \texttt{leeeezy@pku.edu.cn}
}

\begin{document}

\maketitle
\begin{abstract}
This short paper is a fast proof-of-concept that the 3-order B-splines used in Kolmogorov-Arnold Networks (KANs) can be well approximated by Gaussian radial basis functions. Doing so leads to FastKAN, a much faster implementation of KAN which is also a radial basis function (RBF) network. Code available at \url{github.com/ZiyaoLi/fast-kan} .
\end{abstract}

\section{Kolmogorov-Arnold Networks}

Kolmogorov-Arnold Networks\cite{liu2024kan}, also known as KANs, are a class of neural network architectures inspired by the seminal work of Andrey Kolmogorov and Vladimir Arnold on representing continuous functions using a sum of simple functions. KANs are designed to learn and approximate complex, high-dimensional functions efficiently by decomposing them into a series of simpler functions.

The key idea behind KANs is to represent a multivariate function $f(\vx)$ as a superposition of simpler functions, which can be learned more easily by a neural network. Specifically, a KAN represents a function $f(\vx)$ as
\begin{align}
    f(\vx) = \sum_q \Phi_q \left(\sum_p \phi_{q,p}(x_p)\right),
\end{align}
where $\Phi, \phi$ are univariate functions. This hierarchical structure allows KANs to model complex functions by breaking them down into simpler components, which can be learned more efficiently by the network. 

To establish the family of functions of $\Phi$ and $\phi$, \cite{liu2024kan} resorted to B-splines which are theoretically able to approximate well any smooth univariate functions in a given domain. However, during the training stage the variables can shift out of the domain, so an additional rescaling of the spline grids are introduced. Although theoretically founded, these operations of calculating the B-spline basis using the deBoor-Cox iteration \cite{gordon1974b} and rescaling the grids can lead to severe efficiency bottlenecks in KANs.

\section{FastKAN}

This short paper introduces FastKAN, a new implementation of KANs that significantly accelerates the model calculation. Specifically, FastKAN approximates the 3-order B-spline basis using radial basis functions (RBFs) with Gaussian kernels. Also, layer normalization \cite{ba2016layer} are used to prevent the inputs shifting away from the domain of the RBFs. By incorporating these alterations, FastKAN enjoys a much simpler implementation without loss of accuracy.

\section{Gaussian Radial Basis Functions}

Radial basis functions (RBFs) \cite{orr1996introduction,buhmann2000radial} are RBFs are a class of real-valued functions whose value depends solely on the distance from a center point, also known as the radial distance. These basis functions have long been explored in function approximation, machine learning, and pattern recognition tasks. 

The basic idea behind RBFs is to model a function by combining several radially symmetric functions, each centered at a different point in the input space. The output of an RBF network is a linear combination of these radial basis functions, weighted by adjustable coefficients.
Mathematically, an RBF network with $N$ centers can be represented as:
\begin{align}
    f(x) = \sum_{i=1}^N w_i \phi(\|x-c_i\|)
\end{align}
where $w_i$ are the adjustable weights or coefficients and $\phi$ is the radial basis function, which depends on the distance between the input $x$ and a center $x_i$. The Gaussian function is one of the most widely used RBFs, defined as
\begin{align}
    \phi(r) = \exp\left(-\frac{r^2}{2h^2}\right),
\end{align}
where $r$ is the radial distance, and $h$ is a parameter that controls the width or spread of the function.

\begin{figure}[!ht]
\centering
\includegraphics[width=\textwidth]{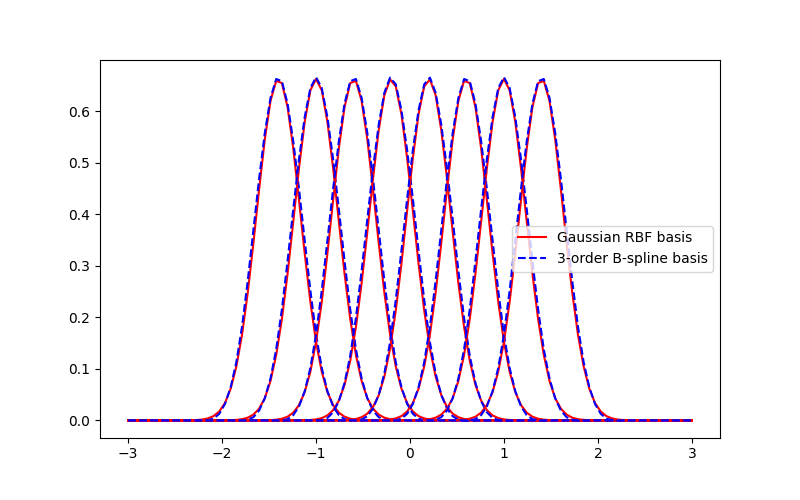}
\caption{Under proper linear transformations, Gaussian RBFs well approximate 3-order B-spline bases with good precision.}
\label{fig}
\end{figure}

Simply by applying linear transformations can one well aligns a series of 3-order B-spline bases to Gaussian radial basis, as is shown in Figure~\ref{fig}. Therefore, one can confidently replaces the B-spline bases calculations with Gaussian RBFs.

\section{Result}

\subsection{Speed}

\newcommand{\efk}{\texttt{efficient\_kan}}

The speed evaluation is conducted over FastKAN and \efk\footnote{https://github.com/Blealtan/efficient-kan}. The latter is a re-implementation of \texttt{pykan} with improved efficiency. Benchmarks are established on NVIDIA V100 GPUs with 32GB memory. To align the performance, \efk\ is configured to use 5 grids with 3-order B-splines, introducing a total of 8 parameters per input; FastKAN also uses 8 centers for the Gaussian RBFs. The test includes a forward calculation a layer of 100 input units and 100 output units, as well as a forward + backward calculation under the same setup. The timing is done by 10 rounds, each round repeats 1000 times of the calculation. No grid rescaling or layer normalization is used during the experiment. Table~\ref{tab} shows the results. FastKAN accelerates the forward speed of \efk\ by 3.33 times.

\begin{table}[]
    \centering
    \begin{tabular}{c|cc|cc}
    \toprule
    Implementation & Fwd. ($\mu$s) & Fwd. acc. & Fwd. + Bwd. ($\mu$s) & Fwd. + Bwd. acc. \\
    \midrule
       \efk & 742$\pm$186 & 1.00& 1160$\pm$18.8 &1.00 \\
        FastKAN & 223$\pm$19 & 3.33& 925$\pm$13.6 &1.25\\
    \bottomrule
    \end{tabular}
    \caption{Forward and backward time.}
    \label{tab}
\end{table}

\subsection{Accuracy}

\begin{figure}[!ht]
\centering
\includegraphics[width=\textwidth]{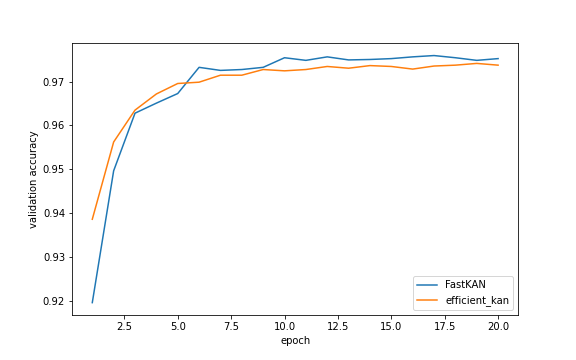}
\caption{Curves of validation accuracy along training on MNIST.}
\label{fig2}
\end{figure}

The accuracy test is conducted on the MNIST dataset\footnote{In fact, MNIST would not be a best case to test KANs. The test is only to show the equivalency of the two models.} provided by \texttt{torchvision}. A total of 20 epochs are trained for the model. For both \efk\ and FastKAN, the models are configured as $[28\times28, 64, 10]$.  Learning curves are shown in Figure~\ref{fig2}. FastKAN is equivalent to (if not better than) KAN.

\section{Discussion}

This short paper propose FastKAN as an acceleration of KAN by replacing the B-splines into Gaussian RBFs. Moreover, this also concludes that KANs are indeed RBF networks with fixed centers.



\bibliographystyle{plain}
\bibliography{ref}

\end{document}